\documentclass[letterpaper, 10 pt, conference]{ieeeconf}  

\IEEEoverridecommandlockouts                  
\usepackage{amssymb}
\usepackage{amsmath}
\usepackage{multirow}
\usepackage[table,xcdraw]{xcolor}
\usepackage{wrapfig}
\usepackage{subfigure}
\usepackage{subcaption}

  {
      \newtheorem{assumption}{Assumption}

  }
\usepackage{hyperref}\hypersetup{colorlinks=true, unicode=true, linkcolor=[rgb]{0.10,0.05,0.67}, citecolor=[rgb]{0.10,0.05,0.67}, filecolor=[rgb]{0.10,0.05,0.67}, urlcolor=[rgb]{0.10,0.05,0.67}}
\usepackage{xcolor}
\usepackage{cleveref}
\usepackage{comment}
\usepackage{graphicx} 
\usepackage{booktabs}
\usepackage{ragged2e}

\usepackage{caption}
\newcommand\rahel[1]{{\color{black} #1}}
\newcommand\melanie[1]{{\color{black} #1}}

\definecolor{lightblue}{rgb}{0.8, 0.9, 1}

\newcommand{\boxedsubcaption}[1]{
\begin{flushleft}
    \begin{minipage}[t]{0.95\linewidth} 
        \fcolorbox{black}{lightblue}{
            \parbox{\linewidth}{\centering \small #1}
        }
    \end{minipage}
\end{flushleft}
}

\newcommand{\boxedsubcaptionsplit}[2]{
\begin{flushleft}
    \begin{minipage}[t]{0.40\linewidth} 
        \fcolorbox{black}{lightblue}{
            \parbox{0.95\linewidth}{\centering \small #1}
        }
    \end{minipage}
    \hspace{0.8em}
    \begin{minipage}[t]{0.40\linewidth} 
        \fcolorbox{black}{lightblue}{
            \parbox{0.95\linewidth}{\centering \small #2}
        }
    \end{minipage}
\end{flushleft}
}

\title{\LARGE \bf
DEMONSTRATE: Zero-shot Language to Robotic Control via Multi-task Demonstration Learning
}

\author{Rahel Rickenbach$^{1}$, Bruce Lee$^{2}$, René Zurbrügg$^{1}$, Carmen Amo Alonso$^{3}$, Melanie N. Zeilinger$^{1}$
\thanks{This work has been submitted to the IEEE for possible publication. Copyright may be transferred without notice, after which this version may no longer be accessible.}
\thanks{This research was partially supported by the ETH AI Center}
\thanks{The authors are with $^{1}$ ETH Zurich, Zurich, Switzerland, $^{2}$ University of Pennsylvania, Philadelphia, USA, and $^{3}$ Stanford University, California, USA. Email correspondence to: {\tt\small \{rrahel,,mzeilinger\}@ethz.ch, brucele@seas.upenn.edu, camoalon@stanford.edu, zrene@leggedrobotics.com.}} %
}

\begin{document}
\maketitle


\begin{abstract}
    The integration of large language models (LLMs) with control systems has demonstrated significant potential in various settings, \rahel{such as task completion with a robotic manipulator}. A main reason for this success is the ability of LLMs to perform in-context learning, which, however, strongly relies on the design of task examples, closely related to the target tasks. Consequently, employing LLMs to formulate optimal control problems often requires task examples that contain explicit mathematical expressions, \melanie{designed by} trained engineers. Furthermore, there is often no principled way to evaluate for hallucination before task execution.
    To address these challenges, we propose \rahel{DEMONSTRATE}, a novel methodology that \rahel{avoids the use of LLMs for} complex optimization problem generations, and instead only relies on the embedding representations of task descriptions. 
    To do this, we leverage tools from inverse optimal control to replace in-context prompt examples with task demonstrations\rahel{, as well as the concept of multitask learning, which ensures target and example task similarity by construction.}  
    Given the fact that hardware demonstrations can easily be collected using teleoperation or guidance of the robot, our approach significantly reduces the reliance on engineering expertise for designing in-context examples. \rahel{Furthermore, the enforced multitask structure enables learning from \melanie{few} demonstrations and assessment of hallucinations prior to task execution.
    We demonstrate the effectiveness of our method through simulation and hardware experiments involving a robotic arm tasked with tabletop manipulation.}
    Videos and code are available at \href{demonstrate-mpc.github.io/}{https://demonstrate-mpc.github.io/}
\end{abstract}



\section{Introduction}

Optimization-based control strategies, such as model predictive control (MPC), have demonstrated great potential in robotics for solving complex tasks in constrained environments. However, their design can be challenging and require extensive domain knowledge. Recent studies have addressed this issue by leveraging large language models (LLMs) to reduce the manual tuning effort and enable human-robot communication through natural language. 
One approach is to directly use a pre-trained LLM to determine low-level robotic actions from a natural language prompt \cite{wang2023prompt,jiang2022vima,Shridhar2022PerceiverActorAM}. This approach typically requires extensive fine-tuning since LLMs often struggle with spatial reasoning tasks \cite{mandi2023roco}. Moreover, it lacks the constraint satisfaction guarantees of advanced control approaches. Another approach is to leverage LLMs \rahel{for} optimal controller design, where the LLM formulates all or some elements of the optimal control problem (OCP), either as code scripts \cite{austin2021program,trivedi2021learning,liang2023code} or mathematical expressions \cite{goyal2019using,yu2023language,ma2023eureka}. While these approaches offer a more principled path toward robotic control via natural language commands, they are hindered by current limitations of LLMs. In particular, given the high sensitivity of LLMs to the prompt, current approaches require highly tuned in-context examples from skilled engineers, which defeats the purpose of using language to ease communication with robots. 
In this paper, we propose \rahel{DEMONSTRATE}, a novel way to interface LLMs and controllers that mitigates this issue. To do so, we build upon the NARRATE pipeline \cite{ismail2024narrate}, where an LLM is used to map a language utterance to a set of objective and constraint functions for an OCP. However, rather than assuming access to highly optimized example prompts, we assume access to a collection of robotic demonstrations for various low-level sub-tasks.  
Using these demonstrations, we form a mapping from embeddings of natural language descriptions for the sub-tasks to the functions describing the OCP. Doing so frees the LLM from the complex task of directly generating the OCP, and provides a quantitative way to assess hallucinations before the task is executed. Moreover, since only embedding representations are required, computation of the mapping output is significantly more efficient than relying on LLM generations.
Hence, this pipeline enables the development and implementation of optimization-based controllers via natural language commands without depending on expert control engineers or their highly optimized contextual example prompts. \rahel{This is achieved via the following two key contributions.}
\begin{itemize}
   \item First, we show how to leverage approaches from the field of inverse reinforcement learning \rahel{(IRL)} and inverse optimal control \rahel{(IOC)} \cite{arora2021}, \rahel{to learn the objective and the constraints from the provided manual demonstrations}.  
   \item Second, we build upon the observation that useful in-context examples are relevant to the task at hand and leverage this idea through the concept of multi-task representation learning \cite{maurer2016benefit,zhang2022multitasksurvey}.
\end{itemize}
In particular, we \rahel{employ multi-task representation learning to make task-similarity, which is often not well-defined and handled by the intuition of an expert, systematic and ensure similarity of target and example tasks by construction. Furthermore, we use} the available demonstrations to extract compressed features of objective and constraint functions and map the embedding space to this shared feature representation. In this way, our architecture can perform zero-shot generalization \rahel{to} new tasks described in natural language.

\section{Related Work}
The use of large language models (LLMs) for control has been successfully applied across various domains, ranging from autonomous driving to robotic manipulation and actuation \cite{wang2023prompt, luu2024context, ismail2024narrate}. To achieve optimal performance, typically one of two strategies are applied: adaptive reasoning or precise prompting. In adaptive reasoning, as demonstrated in \cite{luu2024context}, the LLM is used not only to break down tasks into sub-tasks and design optimization-based controllers but also to provide corrective feedback during execution. Alternatively, precise prompting involves crafting specific, task-relevant prompts, as seen in \cite{ismail2024narrate, wang2023prompt, lee2024affordance}. These prompts can be categorized into two types: visual \cite{lee2024affordance} and textual \cite{ismail2024narrate, wang2023prompt}. However, creating effective textual prompts often requires expertise from skilled engineers or the integration of a well-tuned controller. The quality of these prompts has been shown to significantly impact the performance of LLM-based controllers \cite{brown2020language, zhang2023survey, jiang2020can}. The sensitivity of LLM performance to prompt design has been mitigated through various prompt optimization techniques. For instance, automated methods like autoprompt \cite{shin2020autoprompt, zou2023universal} and reinforcement learning (RL)-based prompt optimization \cite{deng2022rlprompt, zhang2022tempera} have been developed to refine prompts systematically. In cases where prompts are manually designed \cite{wang2023prompt, reynolds2021prompt}, the issue of response sensitivity to user commands has been addressed using uncertainty quantification techniques \cite{park2024}. These methods classify user commands based on their clarity, enabling the LLM to assess the reliability of its responses. Furthermore, such uncertainty quantification can empower the LLM to proactively seek assistance when faced with ambiguous or unclear inputs \cite{ren2023robotsaskhelpuncertainty}. 

\section{Problem Statement}
\label{sec:problemstatement}

In this section, we present the problem of interest and the \rahel{proposed concept, which is then detailed in Section \ref{sec:mticestimation}}. 

\subsection{Problem Setup}
Given a robotic system, the goal is to complete a task $\mathcal{P}$ described by a natural language command, denoted~$\ell_{\mathcal{P}}$. 
A prominent approach in the literature is to convert $\ell_{\mathcal{P}}$ into multiple sub-tasks $\tau$, themselves described with a natural language command $\ell_\tau$, which allow for the translation into an OCP, solved via an MPC \cite{kouvaritakis2016} at each time step~$k$:
\begin{subequations}\label{eqn:generalnmpc}
    \begin{align}
    \underset{\begin{subarray}{c}
      x_0,\ldots,x_N \\ u_0,\ldots,u_{N-1}\end{subarray}}{\text{min}} & \qquad c_\tau(x_0,\ldots,x_N,u_0,\ldots,u_{N-1}) \label{eqn:gnmpc_cost}\\
     s.t. & \qquad x_{i+1} = f(x_i, u_i) ,\quad i=1,\dots N, \label{eqn:gnmpc_dyanimcs} \\
     & \qquad x_0 =  x(k), \\ 
     & \qquad (x_i,u_i) \in \mathcal{X}_\tau,\quad i=1,\dots N.   \label{eqn:gnmpc_constraints}
    \end{align}
    \end{subequations} 
Here, $x(k) \in\mathbb R^{n_x}$ is the state and $u(k) \in\mathbb R^{n_u}$ is the control input, equation \eqref{eqn:gnmpc_dyanimcs} captures the dynamics of the robot, and $c_\tau(\cdot)$ and $\mathcal X_\tau$ represent the cost and constraint set for task $\tau$, respectively. 
While LLMs have shown promising performances in the division of complex tasks into smaller subtasks, in this paper we address the question: \newline
\emph{How can we \rahel{systematically} and reliably map a given natural language utterance $\ell_\tau$ to the OCP representation \rahel{in} \eqref{eqn:generalnmpc} with the appropriate $c_\tau$ and $\mathcal X_\tau$?}

Recent work (NARRATE) \cite{ismail2024narrate} \rahel{addresses} this problem by relying solely on the capabilities of current LLMs to map language utterances describing the task $\tau$ into their corresponding mathematical expressions, $c_\tau$ and $\mathcal X_\tau$. To do this, a language model $L$ takes as input both (i) the language command $\ell_{\tau}$, and (ii) a system prompt $p_{\tau}$, and outputs a symbolic cost function and a set of constraints: \rahel{$(c_{\tau}, \mathcal X_{\tau}) = L(\ell_{\tau}, p_\tau)$}. The system prompt $p_\tau$ is a string of text that consists of $T$ in-context examples that associate a natural language prompt $\ell_{t}$ with the cost $c_{t}$ and constraint set $\mathcal X_{t}$ for each \rahel{example sub-task} $t=1,\dots,T$; i.e., for each example sub-task $t$, the prompt $p_\tau$ contains a 3-tuple  of the form $(\ell_{t}, c_{t}, \mathcal X_{t})$. Three important issues arise with this approach:
\begin{enumerate}
    \item Access to adequate in-context examples requires technical expertise to appropriately design the cost and constraints functions, which is often not available.
    \item Significant \rahel{sensitivity} to the prompt: a slight modification in the prompt can lead to very different \rahel{success rates in execution}.
    \item There is no principled way of assessing hallucinations of the language model before task execution. 
\end{enumerate}

In this paper, we study the problem of how to set up OCP \eqref{eqn:generalnmpc} as specified through natural language commands, \rahel{when} no direct in-context examples are available. Instead, we assume access to \rahel{demonstrations} from either a human user or an existing controller that successfully completes $T$ sub-tasks, each paired with a natural language description $l_{t}$ (with $t=1,\dots,T$). We \rahel{show how} learning from demonstrations of various sub-tasks paired with pre-trained language embedding models \rahel{and enforcing a particular structure of OCP \eqref{eqn:generalnmpc}} bypasses the need for in-context examples of OCPs corresponding to a language description of a sub-task 
and allows for hallucination detection prior to task execution.

\subsection{Approach}
\label{sec:contribution}

\begin{figure*}[]
\vspace{+3mm}
    \centering
    \includegraphics[width=1.0 \textwidth]{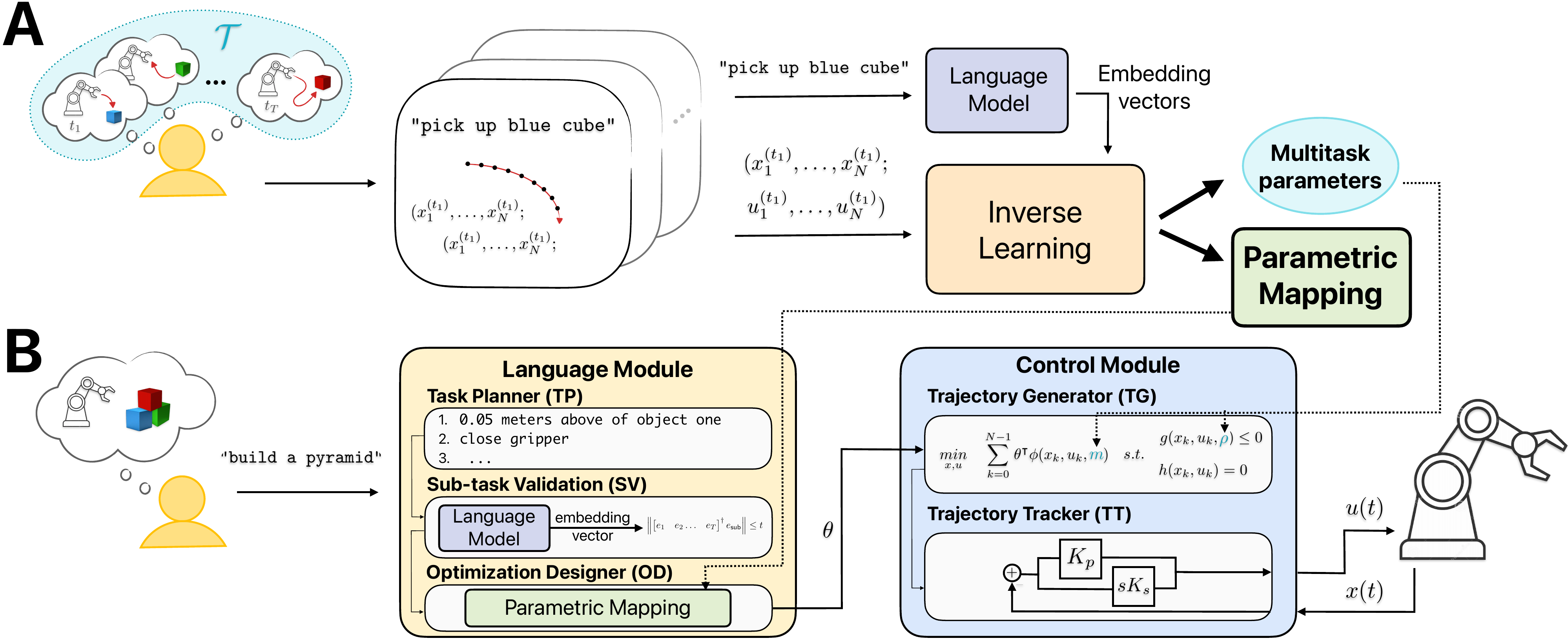}
    \caption{\textbf{A. Offline pipeline:} (a) A user provides sub-task descriptions together with trajectory demonstrations; (b) the LLM computes the embedding vectors for the sub-task descriptions; (c) the demonstrations and the embedding vectors are used to learn (i) a \emph{parametric mapping}, and (ii) the multitask parameters, both used in the online pipeline. \textbf{B. Online pipeline:} Architecture of \cite{ismail2024narrate}, with an added SV Module and modified OD. The LLM is used to find the embedding vectors of the language commands from TP, which are fed into the \emph{parametric map} to compute the feature vector used in the TG block.}
    \vspace{-2em}
    \label{fig:llmpipelinenew}
\end{figure*}
 
To address these challenges, our approach \rahel{relies on two innovations}. First, we leverage tools that allow for the construction of an OCP from successful demonstrations of sub-task examples. Second, \rahel{we systematically ensure that sub-task examples are closely related to potential target sub-tasks, by sharing a common mathematical representation for the OCP. More specifically,} we harness the following concepts.

\paragraph{Learning from Demonstrations} Demonstrations consist of state and action measurements over a trajectory, i.e., $(\mathbf x, \mathbf u):=(x_0,\dots,x_N,u_0,\dots,u_{N-1})$. Given a collection of (potentially suboptimal) demonstrations of an example sub-task $t$, they can be used to learn an objective function $c_{t}$, e.g., with so-called entropy maximization methods \cite{ziebart2008maximum,levine2012continuous}, and the constraints space $\mathcal X_{t}$, via the construction of constraint violating input sequences \mbox{\cite{chou2020,chou2020parametric}.}

\paragraph{Multitask Embedding Mapping} Good choices of in-context example sub-tasks (or demonstrations) carry similarity with potential target tasks. We use this concept in two ways. First, we take advantage of the resulting structure of the language description representation of these sub-tasks in the embedding space. Second, we define a task space $\mathcal S$ such that for all $t\in\mathcal S$, $c_t$ and $\mathcal X_t$ enjoy a specific mathematical structure that can be parameterized via feature vectors \rahel{and enforces structural similarities between example and target sub-tasks.} \newline  
\rahel{DEMONSTRATE proposes an interface between LLMs and optimal controllers that builds on and augments the existing architecture of \cite{ismail2024narrate} for robotic control through natural language commands. The resulting pipeline is shown in Fig.~\ref{fig:llmpipelinenew}. By using a learning-from-demonstrations approach, the user is able to directly provide example sub-task descriptions together with demonstrations, removing the need for technical expertise to create prompt examples in mathematical terms. 
Moreover, the multitask approach allows for enhanced robustness by explicitly leveraging the structure of sub-tasks that share a common task space $\mathcal{S}$ and accordingly, common features describing their representing OCPs. 
By only altering the parametrization of these feature vectors while keeping the remaining structure of the OCP fixed, we ensure a common representation of example and target sub-tasks. Therefore, it becomes possible to directly learn a \textit{parametric mapping} $\mathcal M$ between the embedding vectors of the natural language commands and the parametrization of feature vectors describing the task. 
Hence, generalization to unseen tasks in $\mathcal{S}$ is achieved online via this trained map, alleviating the LLM from complex symbolic function creation.
During generation, the learned \textit{parametric mapping} $\mathcal{M}$ is employed to generate \melanie{new feature combining parameter vectors} of the output sub-task. Consequently, they are guaranteed to fall within the pre-defined task feature space, thus mitigating the risk for LLM hallucinations.}
Additionally, this structure can be used as a correctness assessment by evaluating the \rahel{similarity} between the embedding vector of the generated target and the available example sub-tasks.

\vspace{-0.35em}
\section{Multi-task Demonstration Learning for Language to Control}
\label{sec:mticestimation}

In this section, we provide a detailed description \rahel{for} the different blocks in the proposed pipeline (Fig. \ref{fig:llmpipelinenew}). The components of the offline pipeline are discussed in the first subsection, while the adaptation of the online pipeline from \cite{ismail2024narrate} is discussed in the subsequent subsection.

\subsection{Offline Multitask Learning From Demonstrations}
\label{subsec:offlinepipeline}

The proposed offline pipeline has two main steps: (1) proper design and collection of demonstrations, and (2) learning from collected demonstrations.

\subsubsection{Demonstrations design and data collection} 

\paragraph{Demonstrations design and multitask setting} Inspired by multi-task learning approaches \cite{maurer2016benefit,zhang2022multitasksurvey}, example and target sub-tasks are assumed to share a common mathematical description $\mathcal S$. In particular, for every sub-task $t\in\mathcal S$, we assume the cost $c_t(\cdot)$ and the constraint set $\mathcal X_t$ in \eqref{eqn:gnmpc_cost} and \eqref{eqn:gnmpc_constraints} are of the form:
\begin{subequations}
\begin{align}
    &c_t(x,u) = \theta_t^\intercal \phi(x,u,m_t), \\ 
    &\mathcal{X}_t \subseteq \{x,u \ s.t.\ g_j(x,u,\rho_t) \leq 0, \forall j=1,\hdots,J\},
\end{align}
\label{eqn:mt}
\end{subequations}
for some parameter vectors $\theta_t \in \mathbb{R}^p$, $m_t \in \mathbb{R}^{q_m}$, and $\rho_t \in \mathbb{R}^{q_\rho}$. The maps $g(\cdot), \phi(\cdot) \in C^1$ are design choices and ideally parametrized on the basis of feature vectors \rahel{that capture both target and example sub-tasks. To avoid the necessity of an expert, examples of $\phi(\cdot)$ and $g(\cdot)$  can be pre-implemented in the pipeline, ranging from tailored collections of parametrized norm\melanie{s} and Gaussian Process \cite{lazaro2010sparse} approximating trigonometric functions for $\phi(\cdot)$, to half-spaces and ellipsoids for $g(\cdot)$.} Mathematically, we capture the multitask assumption as follows:
\begin{assumption}\label{assump:multitask}
    All demonstration and target tasks belong to $\mathcal S$ as defined by the cost and constraint set \eqref{eqn:mt}, where $m_{t_1}=m_{t_2}$ and $\rho_{t_1}=\rho_{t_2} \text{ for every pair } t_1,t_2\in\mathcal S$, i.e., all tasks in $\mathcal S$ share the parameter vectors $m$ and $\rho$. \rahel{Task-specificity is defined via $\theta$.}
\end{assumption}

\paragraph{Data Collection} For each demonstrated sub-task example $t_1, \dots, t_T$, we suppose that the demonstrator provides several trajectories, each consisting of a sequence of states and inputs $(\mathbf{x}, \mathbf{u})$, which we assume to be near-optimal, similar to previous work for demonstration learning \cite{levine2012continuous}. Given that the joint estimation of $c_{t_i}(\cdot)$ and $\mathcal X_{t_i}$ is difficult and often ambiguous, we resort to state-of-the-art techniques to carry out cost and constraint estimation independently. Hence, for each \rahel{sub-task example} demonstrations, $t_i$,  we collect $D$ \rahel{demonstrations where the optimal trajectory is not influenced by avoiding the unsafe region $\bar{\mathcal{X}}_t$, i.e., the complement of $\mathcal{X}_t$, as well as $D$ demonstrations where the optimal trajectory is influenced by avoiding the unsafe region $\bar{\mathcal{X}}_t$. While the former demonstrations are indicated with $(\mathbf{\hat{x}}, \mathbf{\hat{u}})_{t_i(d)}$, where $t_i(d)$ with $d\in[1,\dots,D]$ represents the $d^\text{th}$ trajectory sample for task $t_i$ and the hat notation indicates sub-optimality, the later demonstrations are denoted as $(\mathbf{\hat{x}}, \mathbf{\hat{u}})^s_{t_i(d)}$, respectively.}

\subsubsection{Learning from demonstrations}
\label{subsec:learningfromdemonstrations}

\paragraph{Learning of cost function and parametric mapping}
\rahel{This step jointly estimates} the cost parameter $m$ in expression \eqref{eqn:mt} and the \emph{parametric mapping} \rahel{$\mathcal{M}$}, both to be used in the online pipeline. It is comprised of two sub-steps, the \textit{principal component extraction} of embedding vectors and the \textit{learning of the unknown \rahel{cost and mapping} parameters}, which are detailed in the following.  

\textit{Principal component extraction:} The embedding vector representations for the descriptions of the $T$ \rahel{sub-task examples}, satisfying Assumption \ref{assump:multitask}, are computed using a transformer architecture \cite{reimers-2019-sentence-bert, song2020mpnet}. These embeddings, denoted as $e_1, \dots, e_T$, typically reside in a high-dimensional space ($\mathbb{R}^{s}$, where $s \geq 300$). However, since our focus is on the dimensions that exhibit significant variation across sub-tasks, rather than capturing the full semantic content of each sentence, we can often represent them in a much lower-dimensional space. \rahel{Therefore, we compress the embeddings, obtaining $\tilde{e}_1, \dots, \tilde{e}_T$ and reduce the number of demonstrations required to learn the parametric mapping in the subsequent step.}

\textit{Learning of the unknown parameters:} Once the principal component embedding vectors $\tilde e_1,\dots, \tilde e_T$ are computed, the joint estimation problem \rahel{leverages the demonstrations $(\mathbf{\hat{x}}, \mathbf{\hat{u}})_{t(d)}$ and tools from IRL to jointly learn the parametric mapping function $\mathcal{M}$ that maps from the compressed embedding vector $\tilde{e}_t$ to the task specifying feature selection vector $\theta_{t}$ whose structure is pre-determined and parametrized via the parameter vector $q$, i.e., \mbox{$\mathcal{M}:\ q, \tilde e_t \mapsto \theta_t$}, as well as the cost parameter $m$, shared among all tasks in $\mathcal{S}$.}

\paragraph{Learning of constraints}
Provided a cost function $c(\cdot)_t$ has been learnt for sub-task $t$, it is possible to use \rahel{the demonstrations $(\mathbf{\hat{x}}, \mathbf{\hat{u}})^s_{t(d)}$, avoiding an unsafe region $\bar{\mathcal{X}}_t$, to learn its over-approximation and consequently estimate the parameter vector $\rho$ of a parameterized constraint description $g((x, u),\rho) \leq 0$ as in expression \eqref{eqn:mt} that under-approximates the safe region $\mathcal{X}_t$.} 

\subsection{Online Robotic Control } In what follows we briefly describe the components of the proposed online pipeline. 

\subsubsection{Language Module}

\paragraph{Task Planner (TP)} This block receives from the user a natural language command to perform a complex task $\mathcal{P}$. Using a pre-trained LLM, it outputs a list of sub-tasks $\tau_1,\tau_2,\dots$ to be executed sequentially to fulfill the command.
\paragraph{Sub-task Validation} This block ensures that the sub-tasks output by the task planner are sufficiently similar to the example sub-tasks which have been provided by the demonstrator. This comparison is performed by embedding each entry in the list of sub-tasks, and verifying that these embeddings are similar to the embeddings of the example sub-tasks. Let $e_{\mathsf{sub}}$ denote the embedding of a sub-task output by the task planner, and let $e_1, \dots, e_T$ denote embeddings of the demonstration sub-tasks. Then the sub-task validation test verifies that for each sub-task,
\begin{align} \label{eq: subtask validation} \left\| \begin{bmatrix}e_1 & e_2 \dots & e_T \end{bmatrix}^{\dagger} e_{\mathsf{sub}} \right\| \leq t,\end{align}
where $t$ is a user-defined threshold. This check is motivated by characterizations from multi-task learning \cite{zhang2024guarantees} which measure task relatedness via \rahel{the coverage coefficient} between the target task and the space spanned by the source tasks.\footnote{In the multi-task literature, the relatedness would be measured by the distance of the weights on the shared basis $\theta_t$. However, in our setting these weights are output by a neural network mapping the language embeddings to the weights. As this network is trained on the demonstrated sub-tasks, outputs from any embedding tend to be mapped to weights which are similar to the weights for the demonstrated sub-tasks. } 
\paragraph{Optimization Designer (OD)} This block sequentially receives each of the sub-task embedding vectors $e_{\tau_i}$, passing the sub-task validation, and passes it through the precomputed \emph{parametric mapping} to obtain $\theta_{\tau_i}$. 

\subsubsection{Control Module}

\paragraph{Trajectory Generator (TG)} This block receives the parameter vector $\theta_{\tau_i}$ computed by the OD. It solves for an optimal trajectory via an MPC scheme \eqref{eqn:generalnmpc}, with cost and constraint sets structured as in expression \eqref{eqn:mt} and parameters $m$ and $\rho$ extracted from the offline computation. 
\paragraph{Trajectory Tracker (TT)} This block receives the trajectory generated by the TG, and maps it to low-level actions for the given hardware, i.e., torque inputs applied to the motors of a robot, etc. \rahel{Further details are provided in~\cite{ismail2024narrate}.}

\section{\melanie{Details} for Offline Computations}
\label{sec:theoreticaldetails}
\rahel{For the blocks of the offline pipeline, we present solution approaches that are either derived from or build upon existing results in the corresponding domain. Note, however, that depending on the intended application and future developments in the considered field, the presented methods may be adapted or replaced.}

\subsection{Principle Component Analysis of Embedding Vectors}
\label{apx:pca}
\rahel{To achieve the compression of the embedding vectors, we apply Principal Component Analysis (PCA) as follows. 
We stack the embedding vectors of our example tasks into a matrix $E = [e_{1}^{\top}, \hdots, e_{T}^{\top}]^{\top} \in \mathbb{R}^{T \times s}$ and center the data around their column means in $\tilde{E} = E - \mathbf{1}_{s,1}(\frac{1}{T}\sum_{t=1}^{T} e_{t})$, where $\mathbf{1}_{s,1}$ denotes a vector of length $s$ filled with ones. We then compute the singular value decomposition 
\mbox{$\tilde{E} = U\Sigma V^{\top}.$}
Finally, ordering singular values and their corresponding rows and columns of $U$ and $V^{\top}$ in a descending order, a reduction to $z$ principal components is achieved by \mbox{
$\tilde{E}_{\mathrm{PC}} = U_{z} \Sigma_{z}.$}}

\subsection{Inverse Reinforcement Learning for Parameter Estimation}
\label{apx:IRL}
\rahel{To jointly learn the parameter vector $q$ of the parametric mapping function $\mathcal{M}:\ q, \tilde e_t \mapsto \theta_t$, as well as the cost parameter $m$, we propose a formulation that builds upon the entropy maximization approach in \cite{levine2012continuous}, which allows for unknown parameter estimation in a parametrized reward functions $c((\mathbf{x},\mathbf{u});w)$ from $D$ suboptimal demonstrations, indicated with $(\mathbf{\hat{x}}, \mathbf{\hat{u}})_{(d)}$ for $d=1,\dots,D$. The estimation follows from the fact that the probability of obtaining $(\mathbf{\hat{u}})_{(d)}$ for the given $(\hat{x}_0)_{(d)}$ and parameter $w$ is
\begin{align}
\label{eq: expert demonstrations}
    P((\mathbf{\hat{u}})_{(d)}|(\hat{x}_0)_{(d)}; w) = \frac{1}{Z}\exp\left(c((\mathbf{x},\mathbf{u});w)\right)
\end{align}
where we denote with $Z$ the normalization term
\mbox{$Z = \int_{\mathbf{u}} \exp\left(c((\mathbf{x},\mathbf{u});w)\right) d\mathbf{u}.$}
Resorting to the concept of maximum entropy \cite{ziebart2008maximum}, the parameters $w$ are \melanie{defined} to maximize the logarithm of likelihood \eqref{eq: expert demonstrations}, which, as shown in \cite{levine2012continuous}, can be approximated as 
\begin{align*}
    & \log(P((\mathbf{\hat{u}})_{(d)}|(\hat{x}_0)_{(d)}; w)) \\ & \approx \frac{1}{2} g_{(d)}^\top H_{(d)}^{-1} g_{(d)} + \frac{1}{2} \log|-H_{(d)}| - \frac{d_{u}}{2} \log(2\pi), 
\end{align*}
with $g = \nabla_{u} c((\mathbf{\hat{x}}, \mathbf{\hat{u}})_{(d)};w)$ and $H_{i} = \nabla_{u}^2 c((\mathbf{\hat{x}}, \mathbf{\hat{u}})_{(d)};w)$.
Hence, the parameter estimate over all collected demonstrations follows as
\begin{equation}
    \hat{w} = \arg \max_{w} \sum_{d=1}^{D} \frac{1}{2} \left(g_{(d)}\right)^\top H_{(d)}^{-1} g_{(d)} + \frac{1}{2} \log|-H_{(d)}|. \nonumber
\end{equation}
We extend this setup to the multi-task setting for a collection of tasks $t=1,\dots,T$ and sum the likelihood over all available tasks in our task-space $\mathcal{S}$. We leverage the structure in \eqref{eqn:mt} and replace the parameter vector $w$ with the parametric mapping $\mathcal{M}(q,e_t) \rightarrow \theta_t$ as well as the cost parameter $m$. This results in the following optimization problem:
\begin{equation}
\begin{aligned}
    \label{eq: likelihood}
    &\hat q, \hat{m} = \arg \max_{ q, m} \sum_{t=1}^T \sum_{d=1}^{D} \log P((\mathbf{\hat{u}})_{(d)}|(\hat{x}_0)_{(d)}; q,e_t,m), \nonumber
\end{aligned}
\end{equation}
which consequently, learning for a cost instead of a reward function, can be approximated as \begin{align}
    \label{eqn:irl}
    &q, \hat{m}
    = \\ &\arg \min_{ q, m} \sum_{d=1}^D\sum_{t=1}^T  -\frac{1}{2} \left(g_{t(d)}\right)^\top H_{t(d)}^{-1} g_{t(d)} - \frac{1}{2} \log|-H_{t(d)}|, \nonumber
\end{align} 
where $H_{t(d)} := \nabla_{\mathbf{u}}^2 \sum_k \mathcal{M}(q,\tilde e_t)^{\top}\phi((x_k, u_k)_{t(d)}; m)$ and $g_{t(d)} := \nabla_{\mathbf{u}} \sum_k \mathcal{M}(q,\tilde e_t)^{\top}\phi((x_k, u_k)_{t(d)}; m)$.}

\subsection{Constraint Learning from Safe Demonstrations}
\label{apx:constraintlearning}

Following the approach in  \cite{chou2020parametric}, we simulate inputs at random and keep those $(\mathbf{\hat{x}}, \mathbf{\hat{u}})^{us}_{t(d)}$ that satisfy $
c_t((\mathbf{\hat{x}}, \mathbf{\hat{u}})^{us}_{t(d)}) < c_t((\mathbf{\hat{x}}, \mathbf{\hat{u}})^s_{t(d)})$, and hence violate the constraints (as otherwise they would have been attainable by the optimal solution). Obtaining $A$ unsafe simulated trajectories, denoted as $(\mathbf{\hat{x}}, \mathbf{\hat{u}})^{us}_{t(d,a)}$ for $a=1,\dots,A$, the parameter $\rho$ is the solution to 
\begin{subequations}
\begin{align}
    \text{find}& \ \rho\\
    \text{s.t.}& \ g((\hat{x}_k, \hat{u}_k)_{t(d)}^s,\rho) \leq 0, \ \forall k = 1,\hdots,N-1 \label{eqn:const_safe}\\
    & \ g((\hat{x}_k, \hat{u}_k)_{t(d,a)}^{us},\rho) > 0, \ \exists k = 1,\hdots,N-1, \label{eqn:const_unsafe}
    \\
    & \ \forall t = 1,\hdots,T , \ \forall d = 1,\hdots, D, \ \forall a = 1,\hdots, A.
\end{align}
\label{eqn:constraint_learning_concept}
\end{subequations}
\rahel{Estimating an unsafe area $\bar{\mathcal{X}_t}$ that can be described as a boolean conjunction of general convex inequalities, satisfiability modulo convex optimization (SMC) can be employed \cite{shoukry2017smc}. For this purpose, the problem in equation \eqref{eqn:constraint_learning_concept} \melanie{is} reformulated as follows: 
\begin{align*}
    \text{find}& \ \rho,\mathbf{b}^{us},\mathbf{b}^{s}\\
    \text{s.t.}& \ b^{s}_{(d,a,k)} \rightarrow g((\hat{x}_k, \hat{u}_k)^s_{(d)},\rho) \leq 0 \\
    & \ \sum^{N-1}_{k=0}b^{s}_{d,a,k} = N \\
    & \ b^{us}_{(d,a,k)}\rightarrow g((\hat{x}_k, \hat{u}_k)^{us}_{(d,a)},\rho) > 0\\
    & \ \sum^{N-1}_{k=0}b^{us}_{d,a,k} \leq N-1 \\
    & \ \forall k = 0,\hdots,N-1 , \ \forall d = 1,\hdots, D, \ \forall a = 1,\hdots, A.
\end{align*}
Where $\mathbf{b}^{us} = (\mathbf{b}^{us}_{(1,1)},\dots,\mathbf{b}^{us}_{(D,A)}) \in \mathbb{R}^{NAD}$ and $\mathbf{b}^{s} = (\mathbf{b}^{s}_{(1,1)},\dots,\mathbf{b}^{s}_{(D,A)}) \in \mathbb{R}^{NAD}$ define concatenations of the boolean vectors $\mathbf{b}^{us}_{(d,a)} = (b^{us}_{(d,a,0)},\dots,b^{us}_{(d,a,N-1)}) \in \mathbb{R}^N$ and $\mathbf{b}^{s}_{(d,a)} = (b^{s}_{(d,a,0)},\dots,b^{s}_{(d,a,N-1)}) \in \mathbb{R}^N$, which allow for the logic assignment of each convex constraint to an individual binary variable. Such a variable is then evaluated as one, for a fulfilled constraint and zero for a violated constraint, making it possible to indicate constraint violation via their summation and finally find a combination of $\rho$, $\mathbf{b}^{us}$, and $\mathbf{b}^{s}$ that fulfills all the constraints.}

\section{Experiments}
\label{sec:experiments}

The proposed pipeline has been extensively tested in simulation, and its effectiveness has been demonstrated on a real robotic setup.
The chosen environment, as well as the obtained results, are detailed in the following subsections. 

\subsection{Environment}
\label{subsec:environment}
The efficacy of \rahel{DEMONSTRATE} is evaluated using custom simulation environment of \cite{ismail2024narrate}, employing \cite{Andersson2019} and \cite{fiedler2023mpc}, integrated into PandaGym \cite{gallouedec2021panda}. It consists of a seven-axis robotic arm (Franka Panda) on a table. We consider three environments, denoted with ``Cubes'', ``Sponge'', and ``Drawer''. In ``Cubes'', there are four randomly placed cubes of different colors, i.e., \textit{green}, \textit{blue}, \textit{red}, and \textit{orange},  in ``Sponge'', there is a pan placed on a table with a sponge next to it, and in ``Drawer'', there is a cabinet with a drawer set on the table.  ``Sponge'' and ``Drawer'' are used in simulation only, while the ``Cubes'' environment is used for both, simulation and hardware experiments. An illustration of the environments is given in Figure \ref{fig:environments}. The ``Drawer" environment has not been present in the baselines' simulation environment but was added to test the sub-task validation, as well as explore and demonstrate the limits of generalization. It is hence not part of the baseline comparison.

\subsection{Experiment Design}
\label{subsec:experimentdesign}
In this section, we detail the target tasks, designed to test our newly proposed pipeline.

\subsubsection{Target tasks of online pipeline} In all four environments, we execute \emph{pick and place movements} with different objects, belonging to a common task space $\mathcal{S}$. The state of the object and the robot are known throughout the experiment. In the ``Cubes'' environment, we conduct three different experiments. 
Task one is described as ``stack all cubes" and is considered successfully solved if all four cubes are stacked on
top of each other, task two as ``build a pyramid with two cubes at the base and one at the top", and the third task follows as ``write the letter L flat on the table''. In the ``Sponge'' environment the robot arm is tasked to ``grab the sponge and clean the plate with circular movements". Finally, in the ``Drawer'' environment, the robot is asked to ``open the drawer x cm". Each of these tasks is composed of simple sub-tasks that involve moving the gripper some distance and bearing relative to the objects in the scene. The language description for an example of one such sub-task is ``0.05 meters right of object one.''

\begin{table*}
\vspace{+3mm}
\centering
\small

\caption{Simulation Experiments: Comparison of DEMONSTRATE (pre-trained embedding model) 
against NARRATE and VoxPoser on three tasks, 50 repetitions per experiment. We report
the \textit{Success Rate (SR)} for each method and the failure reason categorized
into \textit{Task Planner Failure (TP)}, \textit{Optimization Designer Failure (OD)}, and \textit{Collision (CO)}}
\vspace{-2mm}
\resizebox{0.98\textwidth }{!}{
\begin{tabular}{|l|l|l|l|l|l|l|l|l|}
\hline
Method & \multicolumn{2}{c|}{Stack} & \multicolumn{2}{c|}{L-Shape} & \multicolumn{2}{c|}{Pyramid} & \multicolumn{2}{c|}{Wipe Pan} \\ \cline{2-9}
       & SR [\%] & TP/OD/CO [\%] & SR [\%] & TP/OD/CO [\%] & SR [\%] & TP/OD/CO [\%] & SR [\%] & TP/OD/CO [\%] \\ \hline
\rowcolor[HTML]{D1CFCF}
CaP & \textbf{98} & 2/0/0 & 10 & 16/14/60 & 16 & 76/6/2 & 22 & 48/20/10 \\ \hline
\rowcolor[HTML]{D1CFCF}
VoxPoser & 26 & 0/26/48 & 0 & 0/24/76 & 16 & 22/24/38 & 48 & 22/24/6 \\ \hline
\rowcolor[HTML]{D1CFCF}
NARRATE & 92 & 6/0/2 & 58 & 30/0/12 & 76 & 0/2/22 & 62 & 34/0/4 \\ \hline
DEMONSTRATE & 88 & 6/6/0 & \textbf{60} & 30/2/8 & \textbf{80} & 10/0/10 & \textbf{94} & 2/0/4 \\ \hline
\end{tabular}}
\vspace{-0.8em}
\label{tab:evaluation results}
\end{table*}

\subsubsection{Components of offline pipeline} We use the pipeline with embedding vectors $e_t$ obtained from the ‘all-mpnet-base-v2’ sentence transformer model \cite{song2020mpnet}. The parametric mapping  $\mathcal{M}(q,e_t)$ is chosen as a multi-layer perceptron regressor with four hidden layers of 512 neurons and ReLu activation functions. The embedding $e_t$ is projected onto 20 principal components formed by the sub-task examples. It is trained using 20 demonstrations \rahel{for each of the} 90 different sub-tasks. \rahel{Since the number of required demonstrations increases with increasing suboptimality, we train the framework in simulation, using a demonstrating controller, which completes the sub-tasks examples. The obtained results are intended to serve as a  proof of concept \melanie{for practical settings in which real-world demonstrations would be used.}}

\subsection{Benchmarking Results}
\label{subsec:results}

The performance of DEMONSTRATE, is evaluated for each task, with a total of 50 runs conducted. The results are then compared against those obtained using the existing NARRATE, VoxPoser \cite{huang2023voxposer}, and Code as Policies \cite{liang2023code} architecture. These baseline evaluation results are from \cite{ismail2024narrate}. 
In order to \rahel{highlight} the efficacy of the proposed methodology, we determine the \textit{Success Rate (SR)} and identify the underlying causes of failure. These are categorized as follows:  \textit{Task Planner Failure (TP)}, \textit{Optimization Designer Failure (OD)}, and \textit{Collision (CO)}. A task planner failure occurs when the sub-tasks output by the task planner do not lead to execution of the user command. An \textit{Optimization Designer Failure} occurs if the run was
not successful due to an inaccurate encoding of the sub-task into the optimization problem, meaning that the control objective is not sufficiently accurate to achieve the desired task. In the baselines, optimization designer failures occur due to coding errors resulting in code that is not executable. In DEMONSTRATE, these failures arise due to insufficient precision of the learned cost and constraint functions, \rahel{such as the placement of blocks with insufficient precision on top of each other, resulting in an unstable and collapsing stack.} 
Finally, \textit{Collision} is indicated when the gripper or an object that the gripper is holding collides with another object, resulting in a failed execution of the task. 
The obtained percentages are presented in the \Cref{tab:evaluation results}. \rahel{It can be seen, that DEMONSTRATE shows comparable or higher success rates than the best performing base line model in all 4 tasks.}

\begin{figure}[]
\vspace{+1.5mm}
    \centering 
    \includegraphics[width=0.5\textwidth]{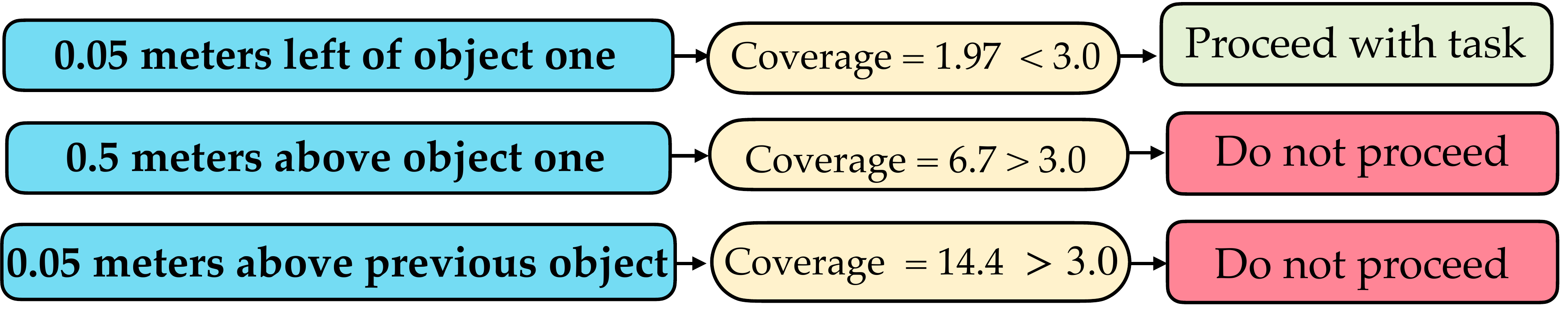}
    \vspace{-5mm}
    \caption{Examples of sub-task validation using the \rahel{coverage coefficient} computed in \eqref{eq: subtask validation} with threshold $t=3.0$. }
    \label{fig:subtask_validation}
    \vspace{-7mm}
\end{figure}

\begin{figure*}[]
\vspace{-3.5mm}
    \centering
    \begin{minipage}{0.67\linewidth} 
    \centering
    \begin{subfigure}
    \raggedright
        \boxedsubcaption{``build a pyramid with the red and green cube as the base and the blue cube at the top"} 
        \vspace{-0.4em}
        \centering
        \includegraphics[width=0.19\textwidth]{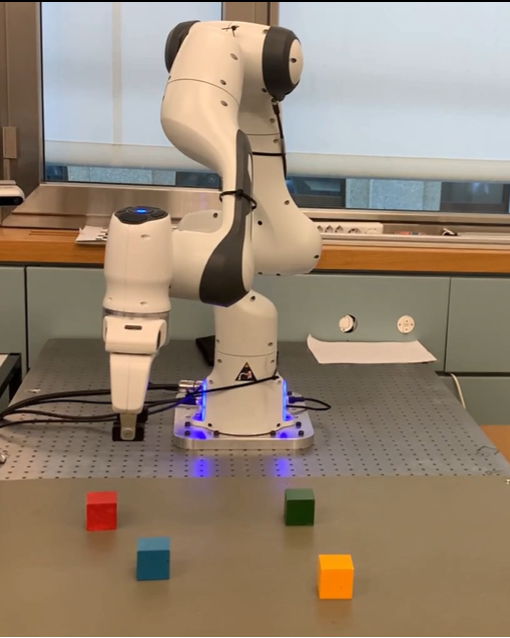}
        \includegraphics[width=0.19\textwidth]{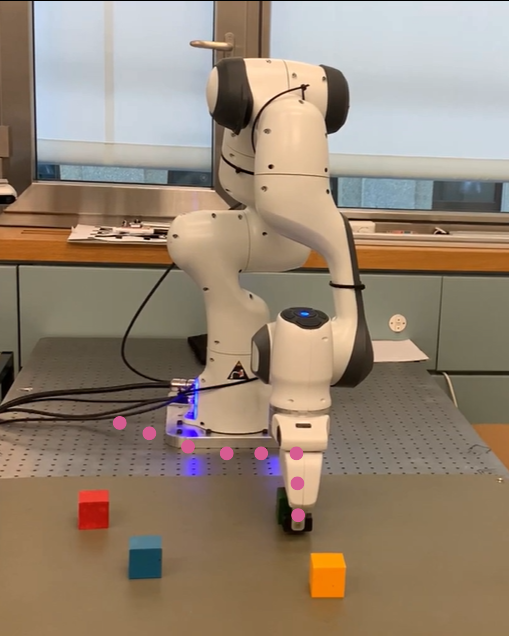}
        \includegraphics[width=0.19\textwidth]{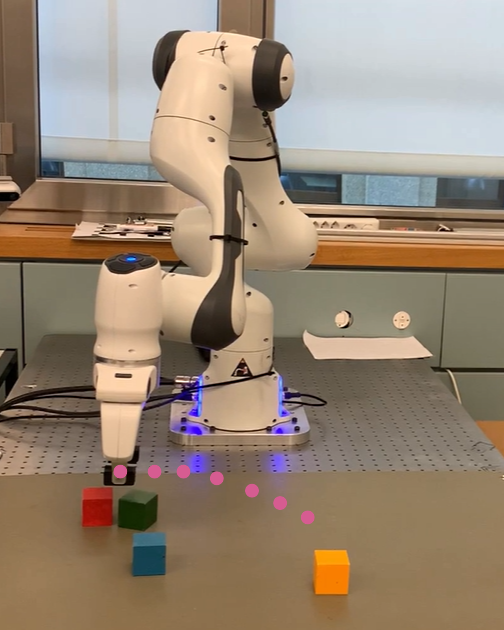}
        \includegraphics[width=0.19\textwidth]{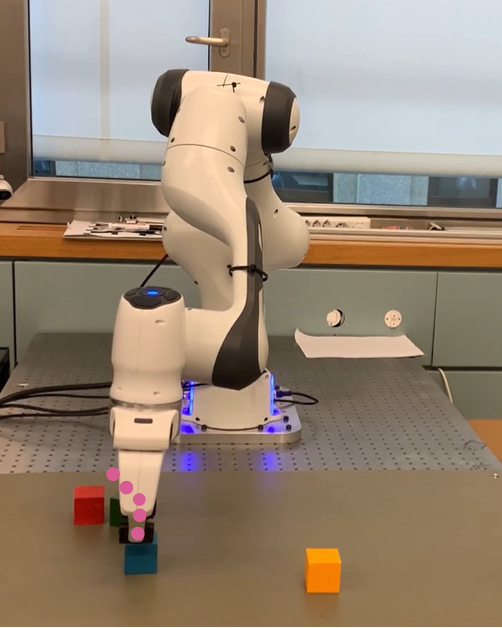}
        \includegraphics[width=0.19\textwidth]{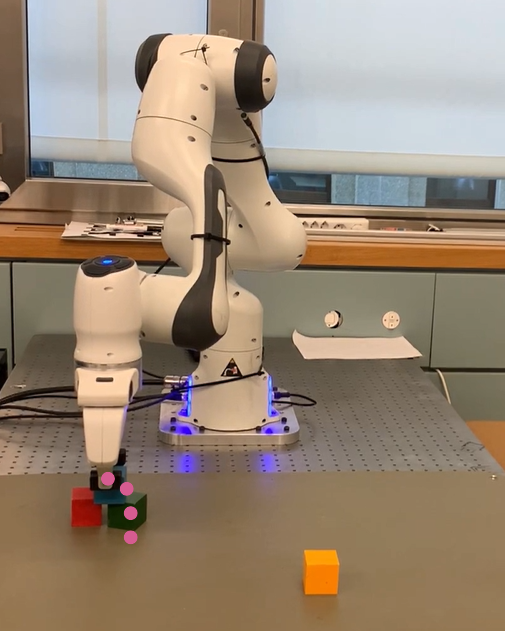}
    \end{subfigure}

    \vspace{-1.8em} 

    \centering
    \begin{subfigure}
        \raggedright
        \boxedsubcaption{``stack all cubes into a tower"} 
        \vspace{-0.4em}
        \centering
        \includegraphics[width=0.19\textwidth]{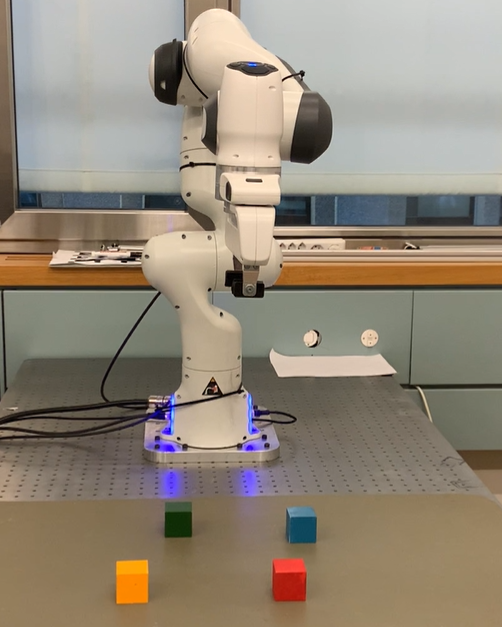}
        \includegraphics[width=0.19\textwidth]{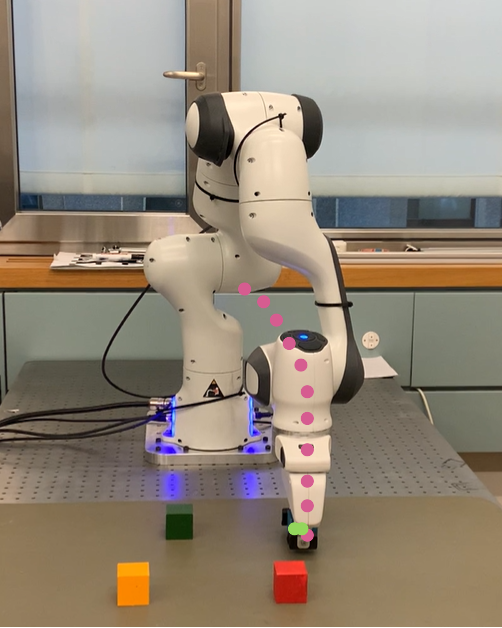}
        \includegraphics[width=0.19\textwidth]{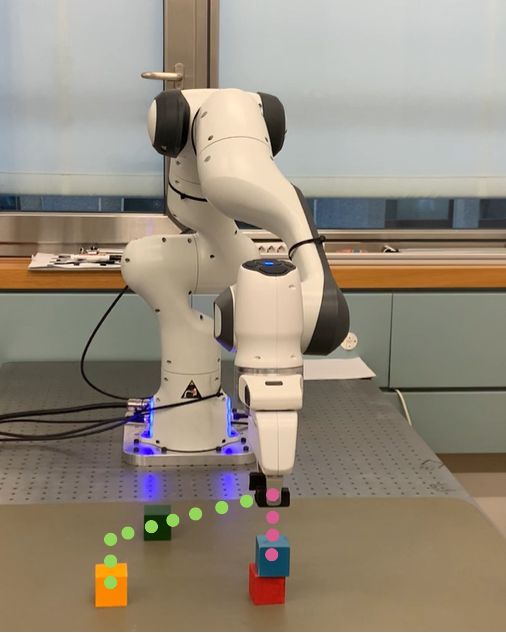}
        \includegraphics[width=0.19\textwidth]{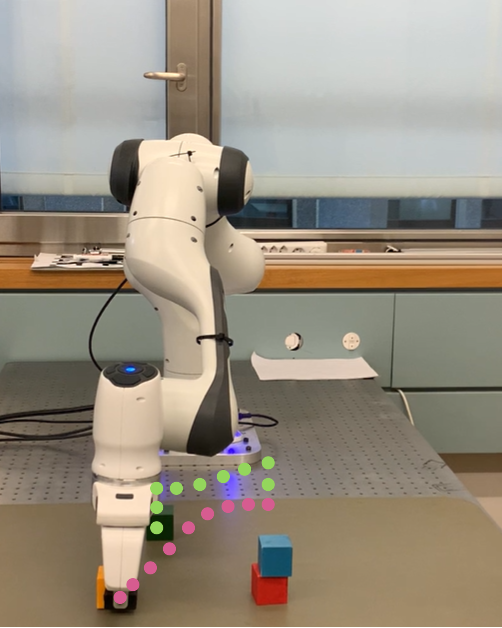}
        \includegraphics[width=0.19\textwidth]{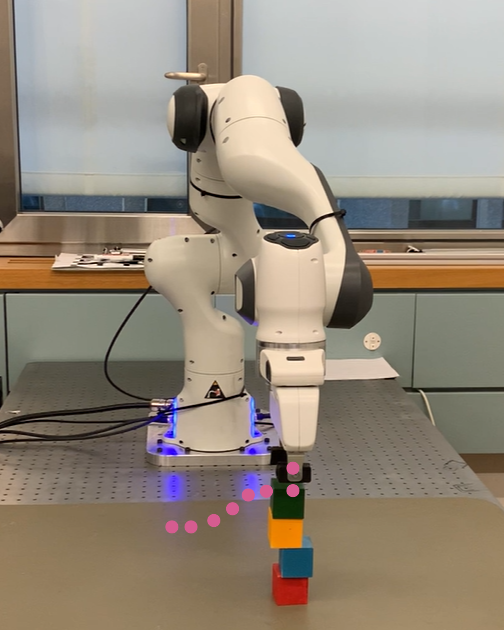}
    \end{subfigure}

    \vspace{-1mm}
    \caption{Visualization of hardware execution to build a pyramid (top) and stack all cubes into a tower (bottom). \rahel{End-effector trajectories are indicated with dotted lines. In case of multiple subsequent trajectories happening in between two pictures, pink trajectories are executed before green trajectories.}}
    \label{fig:example_execution}
    \end{minipage}
    \hfill
   \begin{minipage}{0.30\linewidth} 
   \vspace{-1.0em}
    \begin{subfigure}
        \raggedright
        \boxedsubcaptionsplit{"cubes" sim}{"cubes" world} 
        \vspace{-0.45em}
        \centering
        \hspace{0.1em}
        \includegraphics[width=0.42\textwidth]{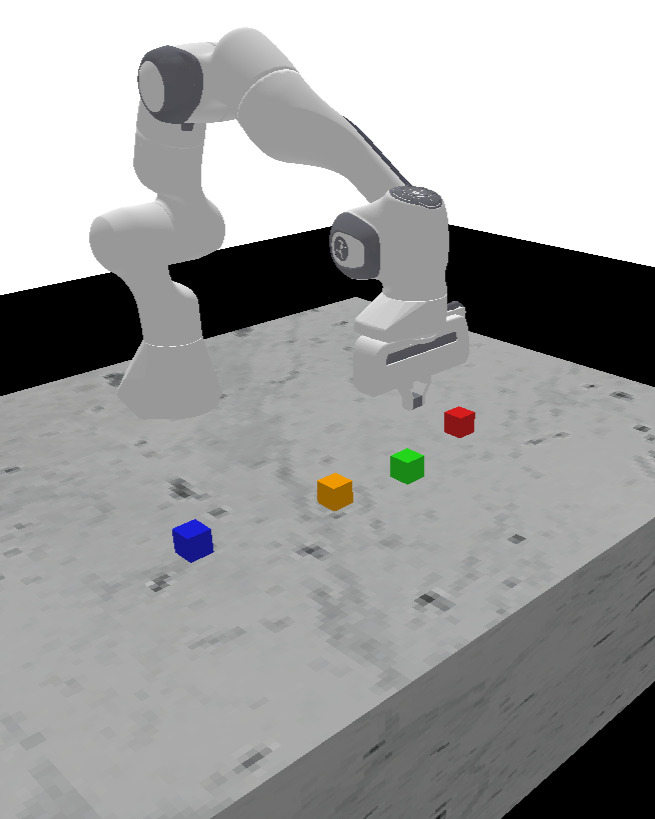}
        \hspace{0.5em}
        \includegraphics[width=0.42\textwidth]{Figures/robot_s1_tw_wp.PNG}
    \end{subfigure}
    \vspace{-0.6em}
    \begin{subfigure}
        \raggedright
        \boxedsubcaptionsplit{"sponge"}{"drawer"} 
        \vspace{-0.45em}
        \centering
        \hspace{0.1em}
        \includegraphics[width=0.42\textwidth]{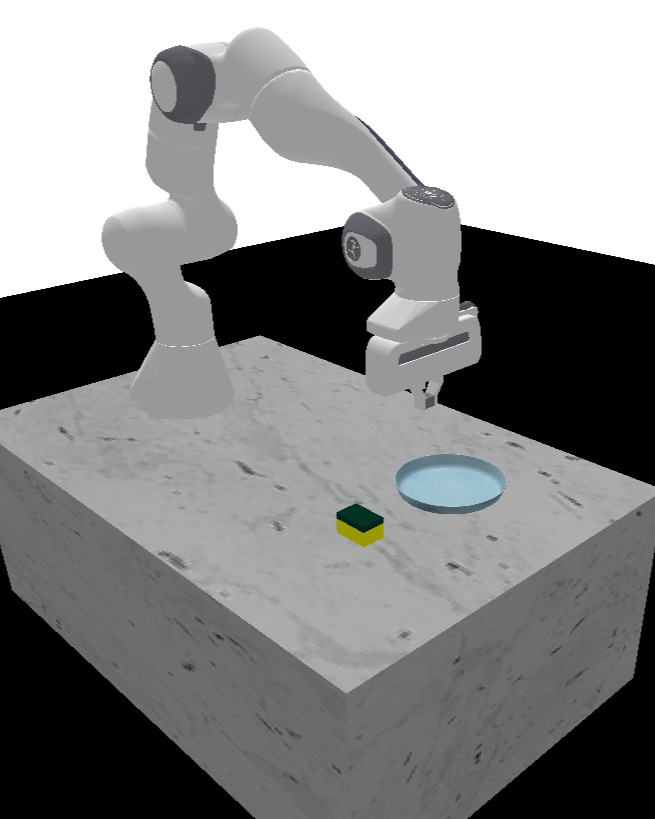}
        \hspace{0.5em}
        \includegraphics[width=0.42\textwidth]{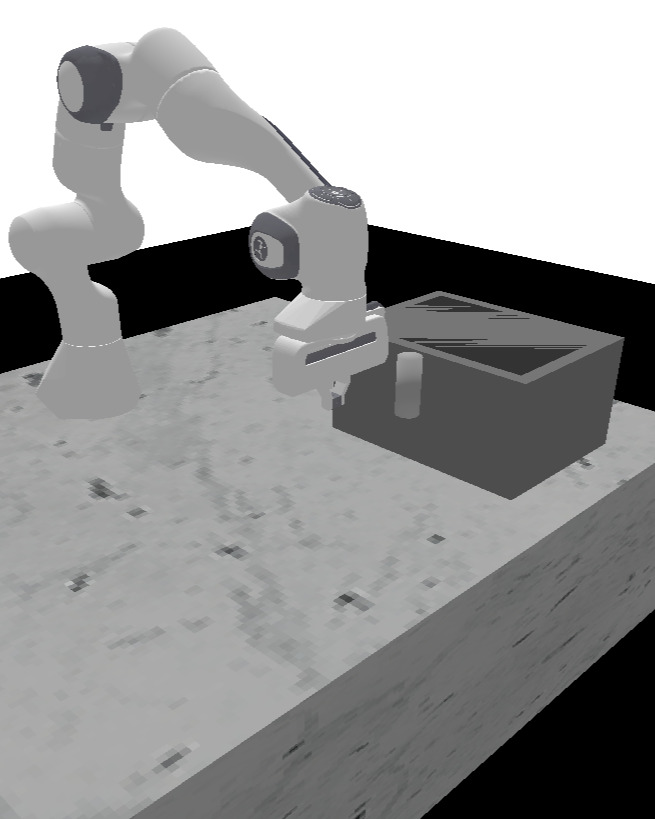}
    \end{subfigure}
    \vspace{-1mm}
    \caption{Illustration of considered environments \rahel{in simulation and real-world experiments}.}
    \label{fig:environments}
    \end{minipage}
    \vspace{-2em}
\end{figure*}

\subsection{Sub-task Validation}
\label{subsec:validation}
 
While the Trajectory Generator (TG) \rahel{outputs} an OCP that aligns with the designed task space $\mathcal{S}$, there are still occasions where the TP suggests sub-tasks that are \rahel{not closely related in their task description with the available sub-task examples}. The sub-task validation procedure of \eqref{eq: subtask validation} is critical to catching these instances. In \Cref{fig:subtask_validation}, we provide illustrative examples where the sub-task validation procedure does and does not interfere, setting the threshold $t$ in \eqref{eq: subtask validation} to $3.0$. In the first instance, the sub-task is of the form provided by the demonstrations, containing a distance along with a maneuver relative to objects one to four. In this case, the sub-task distance is below the threshold. In the second case, a distance \rahel{considerably} larger than those demonstrated is provided, and the third case references the previous sub-task, which is not conform with the provided annotations, both leading to a large \rahel{coverage coefficient}. For the \rahel{proposed DEMONSTRATE architecture} in \Cref{fig:llmpipelinenew}, failure of sub-task validation results in re-planning by the TP, up to a maximum number of tries (set to 5 for the experiments of \Cref{tab:evaluation results}). If the planner fails to generate a list of valid sub-tasks within these trials, the system refuses to attempt the task. In the experiments of \Cref{tab:evaluation results}, the task planner always achieves a valid list of sub-tasks within three attempts. Re-planning was triggered twice for ``Stack'' and  ``Pyramid'', and not at all for ``L'' and  ``Wipe Pan''. 

\subsection{Limits of Generalization}
\label{subsec:generalization}
\begin{table}[h]
\vspace{+2mm}
\centering
\small
\begin{tabular}{|l|l|l|l|}
\hline
``Open the Drawer " &  +``5cm''  &  +``10cm'' &  +``15cm'' \\
\hline 
Result (cm) & 5.8 & 8.9 & Cannot perform \\
\hline
\end{tabular}
\vspace{-1mm}
\caption{Limits: We test the distances which the drawer can be opened via the command ``Open the drawer '' finished by either ``5cm'', ``10cm'', or ``15cm''. We then show the resulting final position for the drawer. }
\vspace{-7mm}
\label{tab:limits of generalization}
\end{table}

\Cref{tab:limits of generalization} shows the results of the procedure applied to the task of opening a drawer, where the demonstration data consisted of sub-tasks moving between 4 and 12 centimeters to the side of an object. The results on the drawer demonstrate that a user command asking the arm to open a drawer to distances close to those demonstrated results in correct movement. \rahel{Asking for larger distances results in the sub-task validation detecting the task to be "not similar enough" to the available sub-task examples} and refusing to perform~it. 

\subsection{Hardware}
The applicability of the proposed approach is demonstrated through real-world experimentation, utilising a Franka Emika Panda robotic arm and two external realsense cameras, as illustrated in Figure \ref{fig:environments}. The effectiveness of the method is evaluated on a subset of the previously introduced tasks, specifically ``Stack", ``L", and ``Pyramid", relying on point cloud estimation to extract the poses of the cubes. A sequence of executed steps for the tasks ``Stack" and ``Pyramid" are visualized in Figure \ref{fig:example_execution}, videos are provided \melanie{in the supplementary material}.

\section{Conclusion}
\label{sec:conclusion}

\textit{Contributions:} We developed a new methodology that makes use of tools derived from inverse reinforcement learning and multitask learning in order to address the challenges associated with prompt design, prompt sensitivity and LLM hallucinations \rahel{in LLM-based controller design}. This is achieved by (i) utilising demonstrations instead of in-context examples consisting of mathematical expressions and (ii) using entropy maximization based inverse reinforcement learning to establish a direct mapping from embeddings of natural language to weights defining the corresponding objective function. This \rahel{avoids the use of the LLM for}
complex objective generations, and instead \rahel{only leverages it to establish} sub-tasks which can be composed to execute the user's command. 

\textit{Results:} The results in tabletop manipulation with a robot arm represent a proof of concept. In particular, they demonstrate that by obtaining a collection of expert demonstrations for a variety of sub-tasks, one can use language-conditioned inverse reinforcement learning with a high-level task planner to execute a diverse array of tasks. The method achieves success rates similar to or better than previous language in controlling approaches on a number of benchmark tasks in \Cref{subsec:results}; however,  it does so at the expense of poorer zero-shot generalization to tasks far from the collected demonstrations, shown in \Cref{subsec:generalization}. To prevent generalization failures, we introduce \emph{Sub-task} validation, tested in \Cref{subsec:validation}, which prevents the execution of tasks too different from those used for training. 

\textit{Limitations:} Building upon state-of-the-art methods from inverse reinforcement learning and constraint learning, the framework currently relies on individual demonstration collection in constrained and unconstrained environments. This is cumbersome and will be addressed in future improvements\rahel{, e.g., by including the constraints in the objective function of the OCP}. Additionally, it was demonstrated in \Cref{subsec:generalization} that the framework does not excel in zero-shot generalization to user commands whose sub-tasks are very different from those \rahel{demonstrated}. This is part of a tradeoff inherent in replacing the LLM used for optimization design with language conditioned inverse reinforcement learning. Improving generalization with the proposed framework would require scaling up the expert demonstrations for a broader range of sub-tasks. 
To move beyond the proof of concept example to a more practically useful framework, one could separate numerical information from semantic information in the sub-task description, and process them separately. This may improve generalization behavior of the type in \Cref{subsec:generalization}. Finally, the framework was tested only in a proof of concept example where it was feasible to quickly collect demonstrations in simulation in a controlled manner. An interesting avenue for future work would be to \rahel{apply} the framework to more sophisticated langauge-enabled robotic control examples, e.g. by using existing datasets of tasks performed by human demonstrators~\cite{o2024open}.

\bibliographystyle{IEEEtran}
\bibliography{IEEEabrv,main}

\begin{thebibliography}{10}
\providecommand{\url}[1]{#1}
\csname url@samestyle\endcsname
\providecommand{\newblock}{\relax}
\providecommand{\bibinfo}[2]{#2}
\providecommand{\BIBentrySTDinterwordspacing}{\spaceskip=0pt\relax}
\providecommand{\BIBentryALTinterwordstretchfactor}{4}
\providecommand{\BIBentryALTinterwordspacing}{\spaceskip=\fontdimen2\font plus
\BIBentryALTinterwordstretchfactor\fontdimen3\font minus \fontdimen4\font\relax}
\providecommand{\BIBforeignlanguage}[2]{{%
\expandafter\ifx\csname l@#1\endcsname\relax
\typeout{** WARNING: IEEEtran.bst: No hyphenation pattern has been}%
\typeout{** loaded for the language `#1'. Using the pattern for}%
\typeout{** the default language instead.}%
\else
\language=\csname l@#1\endcsname
\fi
#2}}
\providecommand{\BIBdecl}{\relax}
\BIBdecl

\bibitem{wang2023prompt}
Y.-J. Wang, B.~Zhang, J.~Chen, and K.~Sreenath, ``Prompt a robot to walk with large language models,'' \emph{arXiv:2309.09969}, 2023.

\bibitem{jiang2022vima}
Y.~Jiang, A.~Gupta \emph{et~al.}, ``Vima: General robot manipulation with multimodal prompts,'' \emph{arXiv:2210.03094}, 2022.

\bibitem{Shridhar2022PerceiverActorAM}
M.~Shridhar, L.~Manuelli, and D.~Fox, ``Perceiver-actor: A multi-task transformer for robotic manipulation,'' in \emph{Conf. on Robot Learning}, 2022.

\bibitem{mandi2023roco}
Z.~Mandi, S.~Jain, and S.~Song, ``Roco: Dialectic multi-robot collaboration with large language models,'' \emph{arXiv:2307.04738}, 2023.

\bibitem{austin2021program}
J.~Austin, A.~Odena \emph{et~al.}, ``Program synthesis with large language models,'' \emph{arXiv:2108.07732}, 2021.

\bibitem{trivedi2021learning}
D.~Trivedi, J.~Zhang \emph{et~al.}, ``Learning to synthesize programs as interpretable and generalizable policies,'' \emph{Adv. Neural Inf. Process. Syst.}, vol.~34, pp. 25\,146--25\,163, 2021.

\bibitem{liang2023code}
J.~Liang, W.~Huang \emph{et~al.}, ``Code as policies: Language model programs for embodied control,'' in \emph{IEEE Int. Conf. on Robot. and Automat. (ICRA)}.\hskip 1em plus 0.5em minus 0.4em\relax IEEE, 2023, pp. 9493--9500.

\bibitem{goyal2019using}
P.~Goyal, S.~Niekum, and R.~J. Mooney, ``Using natural language for reward shaping in reinforcement learning,'' \emph{arXiv:1903.02020}, 2019.

\bibitem{yu2023language}
W.~Yu, N.~Gileadi \emph{et~al.}, ``Language to rewards for robotic skill synthesis,'' \emph{arXiv:2306.08647}, 2023.

\bibitem{ma2023eureka}
Y.~J. Ma, W.~Liang \emph{et~al.}, ``Eureka: Human-level reward design via coding large language models,'' \emph{arXiv:2310.12931}, 2023.

\bibitem{ismail2024narrate}
S.~Ismail, A.~Arbues \emph{et~al.}, ``Narrate: Versatile language architecture for optimal control in robotics,'' in \emph{Int. Conf. Intel. Robots Syst. (IROS)}.\hskip 1em plus 0.5em minus 0.4em\relax IEEE, 2024, pp. 9628--9635.

\bibitem{arora2021}
S.~Arora and P.~Doshi, ``A survey of inverse reinforcement learning,'' \emph{Artif. Intell.}, 2021.

\bibitem{maurer2016benefit}
A.~Maurer, M.~Pontil, and B.~Romera-Paredes, ``The benefit of multitask representation learning,'' \emph{J. Mach. Learn. Res.}, vol.~17, no.~81, pp. 1--32, 2016.

\bibitem{zhang2022multitasksurvey}
Y.~Zhang and Q.~Yang, ``A survey on multi-task learning,'' \emph{IEEE Trans. Knowl. Data Eng.}, vol.~34, no.~12, pp. 5586--5609, 2022.

\bibitem{luu2024context}
Q.~K. Luu, X.~Deng \emph{et~al.}, ``Context-aware llm-based safe control against latent risks,'' \emph{arXiv:2403.11863}, 2024.

\bibitem{lee2024affordance}
O.~Y. Lee, A.~Xie \emph{et~al.}, ``Affordance-guided reinforcement learning via visual prompting,'' \emph{arXiv:2407.10341}, 2024.

\bibitem{brown2020language}
T.~Brown, B.~Mann \emph{et~al.}, ``Language models are few-shot learners,'' \emph{Adv. Neural Inf. Proc. Syst.}, vol.~33, pp. 1877--1901, 2020.

\bibitem{zhang2023survey}
H.~Zhang, H.~Song \emph{et~al.}, ``A survey of controllable text generation using transformer-based pre-trained language models,'' \emph{ACM Computing Surveys}, vol.~56, no.~3, pp. 1--37, 2023.

\bibitem{jiang2020can}
Z.~Jiang, F.~F. Xu \emph{et~al.}, ``How can we know what language models know?'' \emph{Trans. Assoc. Comput. Linguistics}, vol.~8, pp. 423--438, 2020.

\bibitem{shin2020autoprompt}
T.~Shin, Y.~Razeghi \emph{et~al.}, ``Autoprompt: Eliciting knowledge from language models with automatically generated prompts,'' \emph{arXiv:2010.15980}, 2020.

\bibitem{zou2023universal}
A.~Zou, Z.~Wang \emph{et~al.}, ``Universal and transferable adversarial attacks on aligned language models,'' \emph{arXiv:2307.15043}, 2023.

\bibitem{deng2022rlprompt}
M.~Deng, J.~Wang \emph{et~al.}, ``Rlprompt: Optimizing discrete text prompts with reinforcement learning,'' \emph{arXiv:2205.12548}, 2022.

\bibitem{zhang2022tempera}
T.~Zhang, X.~Wang \emph{et~al.}, ``Tempera: Test-time prompting via reinforcement learning,'' \emph{arXiv:2211.11890}, 2022.

\bibitem{reynolds2021prompt}
L.~Reynolds and K.~McDonell, ``Prompt programming for large language models: Beyond the few-shot paradigm,'' in \emph{Extended abstracts of the 2021 CHI conference on human factors in computing systems}, 2021, pp. 1--7.

\bibitem{park2024}
J.~Park, S.~Lim \emph{et~al.}, ``Clara: Classifying and disambiguating user commands for reliable interactive robotic agents,'' \emph{IEEE Robot. Automat. Lett.}, vol.~9, no.~2, pp. 1059--1066, 2024.

\bibitem{ren2023robotsaskhelpuncertainty}
A.~Z. Ren, A.~Dixit, A.~Bodrova, S.~Singh, S.~Tu, N.~Brown, P.~Xu, L.~Takayama, F.~Xia, J.~Varley, Z.~Xu, D.~Sadigh, A.~Zeng, and A.~Majumdar, ``Robots that ask for help: Uncertainty alignment for large language model planners,'' 2023.

\bibitem{kouvaritakis2016}
B.~Kouvaritakis and M.~Cannon, ``Model predictive control,'' \emph{Switzerland: Springer Int. Publishing}, 2016.

\bibitem{ziebart2008maximum}
B.~D. Ziebart, A.~L. Maas \emph{et~al.}, ``Maximum entropy inverse reinforcement learning,'' in \emph{Conf. Artif. Intel.}, vol.~8.\hskip 1em plus 0.5em minus 0.4em\relax Chicago, IL, USA, 2008, pp. 1433--1438.

\bibitem{levine2012continuous}
S.~Levine and V.~Koltun, ``Continuous inverse optimal control with locally optimal examples,'' in \emph{Proc. Int. Conf. Mach. Learn.}, 2012, pp. 475--482.

\bibitem{chou2020}
G.~Chou, D.~Berenson, and N.~Ozay, ``Learning constraints from demonstrations,'' in \emph{Algorithmic Found. of Robot. XIII}.\hskip 1em plus 0.5em minus 0.4em\relax Springer, 2020, pp. 228--245.

\bibitem{chou2020parametric}
G.~Chou, N.~Ozay, and D.~Berenson, ``Learning parametric constraints in high dimensions from demonstrations,'' in \emph{Conf. Robot Learn.}\hskip 1em plus 0.5em minus 0.4em\relax PMLR, 2020, pp. 1211--1230.

\bibitem{lazaro2010sparse}
M.~L{\'a}zaro-Gredilla, J.~Quinonero-Candela \emph{et~al.}, ``Sparse spectrum gaussian process regression,'' \emph{J. Mach. Learn. Res.}, vol.~11, pp. 1865--1881, 2010.

\bibitem{reimers-2019-sentence-bert}
N.~Reimers and I.~Gurevych, ``Sentence-bert: Sentence embeddings using siamese bert-networks,'' in \emph{Proc Conf. Empirical Methods in Natural Language Processing}.\hskip 1em plus 0.5em minus 0.4em\relax Association for Computational Linguistics, 11 2019.

\bibitem{song2020mpnet}
K.~Song, X.~Tan \emph{et~al.}, ``Mpnet: Masked and permuted pre-training for language understanding,'' \emph{Adv. Neural Inf. Process. Syst.}, vol.~33, pp. 16\,857--16\,867, 2020.

\bibitem{zhang2024guarantees}
T.~T. Zhang, B.~D. Lee \emph{et~al.}, ``Guarantees for nonlinear representation learning: non-identical covariates, dependent data, fewer samples,'' \emph{arXiv:2410.11227}, 2024.

\bibitem{shoukry2017smc}
Y.~Shoukry, P.~Nuzzo, A.~L. Sangiovanni-Vincentelli, S.~A. Seshia, G.~J. Pappas, and P.~Tabuada, ``Smc: Satisfiability modulo convex optimization,'' in \emph{Proceedings of the 20th international conference on hybrid systems: Computation and control}, 2017, pp. 19--28.

\bibitem{Andersson2019}
J.~A.~E. Andersson, J.~Gillis, G.~Horn, J.~B. Rawlings, and M.~Diehl, ``{CasADi} -- {A} software framework for nonlinear optimization and optimal control,'' \emph{Mathematical Programming Computation}, 2019.

\bibitem{fiedler2023mpc}
F.~Fiedler, B.~Karg, L.~L{\"u}ken, D.~Brandner, M.~Heinlein, F.~Brabender, and S.~Lucia, ``do-mpc: Towards fair nonlinear and robust model predictive control,'' \emph{Control Engineering Practice}, 2023.

\bibitem{gallouedec2021panda}
Q.~Gallou{\'e}dec, N.~Cazin \emph{et~al.}, ``panda-gym: Open-source goal-conditioned environments for robotic learning,'' \emph{arXiv:2106.13687}, 2021.

\bibitem{huang2023voxposer}
W.~Huang, C.~Wang \emph{et~al.}, ``Voxposer: Composable 3d value maps for robotic manipulation with language models,'' \emph{arXiv:2307.05973}, 2023.

\bibitem{o2024open}
A.~O’Neill, A.~Rehman, A.~Maddukuri, A.~Gupta, A.~Padalkar, A.~Lee, A.~Pooley, A.~Gupta, A.~Mandlekar, A.~Jain \emph{et~al.}, ``Open x-embodiment: Robotic learning datasets and rt-x models: Open x-embodiment collaboration 0,'' in \emph{Int. Conf. Robot. Automation (ICRA)}.\hskip 1em plus 0.5em minus 0.4em\relax IEEE, 2024, pp. 6892--6903.

\end{thebibliography}

\newpage


\clearpage


\end{document}